\documentclass{article}

\PassOptionsToPackage{numbers, sort&compress}{natbib}
\usepackage[main, final]{neurips_2026}

\usepackage[utf8]{inputenc} 
\usepackage[T1]{fontenc}    
\usepackage{hyperref}       
\usepackage{url}            
\usepackage{booktabs}       
\usepackage{amsfonts}       
\usepackage{nicefrac}       
\usepackage{microtype}      
\usepackage[table]{xcolor}   
\usepackage{graphicx}
\usepackage{amsmath}
\usepackage{cleveref}
\usepackage{booktabs}
\usepackage{multirow}
\usepackage{algorithm}
\usepackage{algorithmicx, algpseudocode}
\usepackage{placeins}
\usepackage{float}

\crefname{table}{Table}{Tables}
\Crefname{table}{Table}{Tables}
\crefname{section}{Section}{Sections}
\Crefname{section}{Section}{Sections}
\crefname{figure}{Figure}{Figures}
\Crefname{figure}{Figure}{Figures}

\title{Optimized Minimal 4D Gaussian Splatting for Efficient Dynamic Scene Representation}

\author{%
\textbf{Minseo Lee}$^1$\thanks{Equal contribution} \quad
\textbf{Byeonghyeon Lee}$^1$\footnotemark[1] \quad
\textbf{Lucas Yunkyu Lee}$^{2,3}$ \quad
\textbf{Eunsoo Lee}$^4$ \\
\textbf{JunYoung Jeong}$^5$ \quad
\textbf{Sangmin Kim}$^2$ \quad
\textbf{Seunghyeon Song}$^4$ \quad
\textbf{Joo Chan Lee}$^4$ \\
\textbf{Jong Hwan Ko}$^4$ \quad
\textbf{Jaesik Park}$^2$ \quad
\textbf{Eunbyung Park}$^1$\thanks{Corresponding author}%
\\[2ex]
\normalfont 
$^1$ Yonsei University \quad
$^2$ Seoul National University \quad
$^3$ POSTECH \quad
$^4$ Sungkyunkwan University \\
$^5$ Electronics and Telecommunications Research Institute \\
{\small\url{https://minshirley.github.io/OMG4/}}
}

\begin{document}

\maketitle

\begin{figure}[h]
    \centering
    \includegraphics[width=1.0\linewidth]{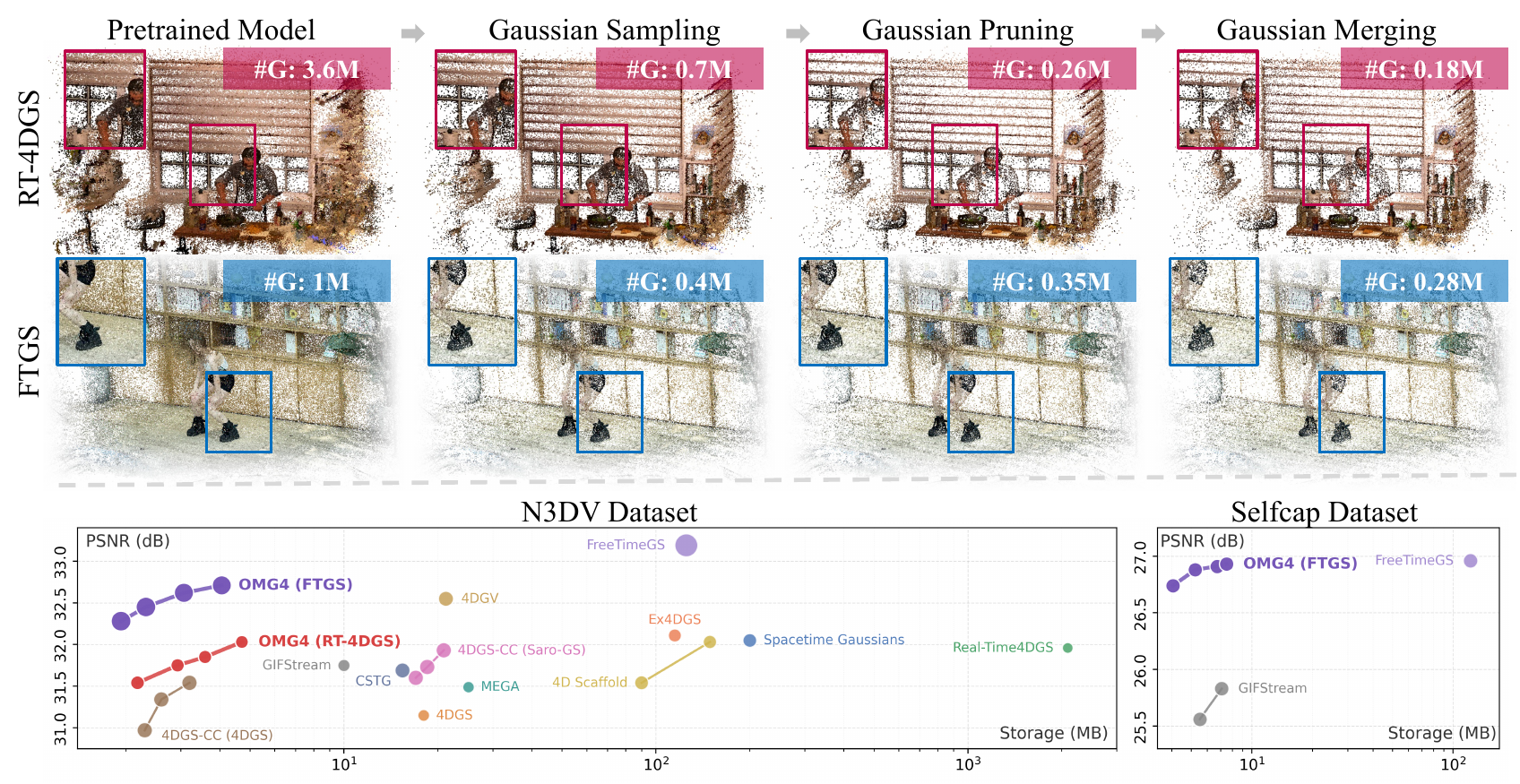}
    \vspace{-0.7cm}
    \caption{The overall \textit{OMG4} pipeline and performance comparison. Top: Multi-stage compression applied to Real-Time4DGS (RT-4DGS) on \textit{Cook Spinach} (N3DV dataset) and FreeTimeGS (FTGS) on \textit{Dance1} (Selfcap dataset). Bottom: Rate-distortion comparison across 4DGS methods (larger circles indicate higher FPS).}
    \label{fig:teaser}
\end{figure}

\begin{abstract}
4D Gaussian Splatting has emerged as a new paradigm for dynamic scene representation, enabling real-time rendering of scenes with complex motions. However, it faces a major challenge of storage overhead, as millions of Gaussians are required for high-fidelity reconstruction. 
Compressing explicit 4D Gaussians is challenging because they exhibit heterogeneous redundancy across static and dynamic regions, and their temporal attributes make direct quantization unstable.
In this work, we present \textit{OMG4} (Optimized Minimal 4D Gaussian Splatting), a framework that constructs a compact set of salient Gaussians capable of faithfully representing 4D Gaussian models.
Our method progressively reduces Gaussians in three stages: (1) Gaussian Sampling to identify primitives critical to reconstruction fidelity, (2) Gaussian Pruning to remove redundancies, and (3) Gaussian Merging to fuse primitives with similar characteristics.
In addition, we integrate implicit appearance compression and extend Sub-Vector Quantization (SVQ) to 4D representations with a staged quantization scheme, further reducing storage while preserving quality.
Extensive experiments on standard benchmark datasets demonstrate that \textit{OMG4} achieves a favorable rate-distortion trade-off over recent state-of-the-art methods, reducing model sizes by over 60\% while maintaining reconstruction quality.
These results demonstrate the effectiveness of \textit{OMG4} as a practical framework for compact and high-fidelity 4D scene representation.
\end{abstract}

\section{Introduction}
\label{sec:intro}

3D Gaussian Splatting~\cite{3DGS} has recently achieved remarkable success, becoming a backbone for diverse 3D vision tasks ranging from 3D novel view synthesis to downstream applications such as visual odometry~\cite{keetha2024splatam,yan2024gs}, 3D scene editing~\cite{vcedit,editsplat}, and degradation-aware rendering~\cite{lee2024deblurring,yu2024mip,wan2025s2gaussian}. Building on this success, 4D Gaussian representations that explicitly model space and time have emerged as a new paradigm for dynamic scene reconstruction~\cite{realtime-4dgs,4dgs,deformabel3dgs,freetimegs}. 
By augmenting Gaussian primitives with temporal parameters, these approaches can capture object motion and appearance variations over time, enabling photorealistic real-time novel view synthesis for applications such as free-viewpoint video~\cite{QUEEN,GIFStream,sun20243dgstreamontheflytraining3d}, autonomous driving simulation~\cite{autosplat,driving}, and VR/AR~\cite{pan2025telegs,jiang2024vr,xu2023vr}.

Existing 4DGS methods typically follow either deformation-based modeling from a canonical 3D Gaussian set~\cite{4dgs,deformabel3dgs} or direct optimization of 4D Gaussians in a unified space-time volume~\cite{realtime-4dgs}. 
The latter often achieves high fidelity for complex object motions, but remains storage-heavy because millions of primitives with high-dimensional and time-varying attributes are required (e.g., Real-Time4DGS~\cite{realtime-4dgs} produces millions of 4D Gaussians, consuming over a gigabyte of memory). This overhead limits practicality in real-time, mobile, and streaming scenarios, highlighting the need for effective compression techniques that preserve both visual fidelity and rendering speed.

Compressing explicit 4D Gaussian representations, however, is not a straightforward extension of static 3DGS compression. 4D Gaussian exhibits heterogeneous redundancy across static and dynamic regions, and introduces temporal attributes that are tightly coupled with spatial geometry and appearance, hence directly quantizing all attributes at once leads to unstable optimization. These observations suggest that effective 4DGS compression should jointly address primitive-level redundancy and stable spatio-temporal attribute compression.

Several works have attempted to alleviate the significant storage requirement of explicit 4D representations~\cite{1000+,MEGA,GIFStream}. 4DGS-1K~\cite{1000+} presents a lifespan-based importance score to prune short-lived Gaussians, reducing the number of primitives and compressing the representation to hundreds of megabytes. 
GIFStream~\cite{GIFStream} performs motion-aware pruning using feature streams, while Light4GS~\cite{Light4GS} and ADC-GS~\cite{ADC-GS} adopt context modeling or anchor-based structures to compress deformation-based dynamic Gaussians~\cite{4dgs}. 
Despite these advances, existing methods still require tens of megabytes to represent only a few seconds of dynamic content, limiting their practicality.

In this paper, we propose \textit{OMG4} (Optimized Minimal 4D Gaussian Splatting), a framework for compact and high-fidelity dynamic scene representation. 
Starting from a pretrained Real-Time4DGS~\cite{realtime-4dgs}, \textit{OMG4} progressively constructs a compact primitive set through Gaussian Sampling, Gaussian Pruning, and Gaussian Merging, which identify static-dynamic salient primitives, remove redundant ones, and fuse similar primitives in local space-time neighborhoods.
Given the resulting compact set, we further compress Gaussian attributes by integrating implicit appearance modeling and extending Sub-Vector Quantization (SVQ)~\cite{OMG} to 4D representations. Importantly, we show that attribute compression becomes tractable only after Gaussian reduction, and introduce a staged 3D-to-4D SVQ scheme that mitigates the instability of simultaneously quantizing spatial and temporal attributes.

We conduct experiments on diverse benchmarks, including N3DV~\cite{n3dv}, ENeRF-Outdoor~\cite{enerf}, and SelfCap~\cite{selfcap}. To the best of our knowledge, we achieve state-of-the-art performance, significantly reducing the volume of the baseline model by three orders of magnitude. Notably, compared to GIFStream~\cite{GIFStream}, a recent state-of-the-art approach, our method reduces storage size by approximately 64\% (from 10.0 MB to 3.59 MB in the N3DV dataset). We further demonstrate the generality of \textit{OMG4} by applying it to other dynamic scene representations, including STG~\cite{spacetimegaussian} and FreeTimeGS~\cite{freetimegs}. 

To sum up, our contributions are as follows:
\begin{itemize}
\item We propose a novel static-dynamic Gaussian minimization framework, progressively reducing the number of Gaussians through Gaussian Sampling, Gaussian Pruning, and Gaussian Merging, while maintaining the reconstruction quality.
\item We extend Sub-Vector Quantization (SVQ) to dynamic 4D representations with a staged quantization scheme, and empirically show that such attribute compression becomes tractable only after sufficient Gaussian reduction via our pipeline.
\item We validate the generality of our method by applying it to various dynamic scene representation baselines, including Real-Time4DGS~\cite{realtime-4dgs}, STG~\cite{spacetimegaussian}, and FreeTimeGS~\cite{freetimegs}.
\item To the best of our knowledge, we achieve a state-of-the-art rate-distortion trade-off, reducing the baseline model from gigabytes to a few megabytes.
\end{itemize}

\section{Related Work}
\subsection{3D Gaussian Splatting and Compression}
3D Gaussian Splatting (3DGS)~\cite{3DGS} enables real-time and high-fidelity novel view synthesis with explicit Gaussian primitives, but its large number of primitives leads to substantial storage overhead. 
Recent 3DGS compression methods reduce this cost through pruning-based strategies~\cite{lightgaussian,lp3dgs}, learnable masks and vector quantization~\cite{compact3dgaussian}, sub-vector or locality-aware attribute encoding~\cite{OMG,locogs}, and structured anchor-based representations~\cite{scaffoldgs,hac}. Specifically, OMG~\cite{OMG} introduces sub-vector quantization (SVQ) that splits the vectors into multiple small sub-vectors and applies VQ, achieving significant performance improvement. Although these approaches effectively reduce the storage requirements of 3DGS, they do not straightforwardly generalize to 4D representations.

\subsection{Dynamic 3D Gaussian Splatting and Compression}
Early efforts to extend static scene reconstruction to dynamic settings~\cite{Hexplane,K-Planes,mixedneuralvoxels,D-NeRF,DyNeRF} were primarily based on neural volumetric rendering, but suffered from high computational costs. More recently, many studies have sought to extend 3DGS to dynamic scenes, which can be broadly categorized into two approaches: (1) representing dynamic scenes with 4D Gaussian primitives that jointly encode spatial and temporal dimensions~\cite{realtime-4dgs,spacetimegaussian}, or (2) deforming 3D Gaussians at each timestamp via a deformation field~\cite{DynMF,embedding-based-deformation,4dgs,deformabel3dgs}. Real-Time4DGS~\cite{realtime-4dgs} achieves high-fidelity dynamic reconstruction by parameterizing the 4D volume with a set of 4D Gaussians. Recently, FreeTimeGS~\cite{freetimegs} has shown promising performance by moving 3D Gaussians over time, leveraging motion vectors.

Similar to 3DGS, dynamic extensions also require a significant number of primitives, motivating the development of lightweight methods. Within the first category, which employs 4D Gaussian primitives, CSTG~\cite{CSTG}, MEGA~\cite{MEGA}, and 4DGS-1K~\cite{1000+} have been proposed. Among them, 4DGS-1K~\cite{1000+} reduces storage by discarding Gaussians with short lifespans, achieving a model size of around 50 MB after post-processing. In contrast, other works~\cite{Light4GS,ADC-GS,4DGS-CC,GIFStream} focus on compressing deformable 3DGS.
In this paper, we primarily target Real-Time4DGS~\cite{realtime-4dgs}, by jointly reducing spatio-temporal primitive redundancy and compressing Gaussian attributes.

\section{Preliminary}
Our framework is designed to be applicable to various Gaussian-based dynamic scene representations, including Real-Time4DGS~\cite{realtime-4dgs}, FreeTimeGS (FTGS)~\cite{freetimegs}, and Spacetime Gaussians (STG)~\cite{spacetimegaussian}. Despite differences in their temporal parameterizations, these methods share an explicit representation of dynamic scenes using Gaussian primitives.

Real-Time4DGS represents a dynamic scene with a set of 4D Gaussian primitives defined by a 4D mean at spatio-temporal space, an anisotropic 4D covariance, an opacity, and spherical harmonics (SH) coefficients. During rendering, 4D Gaussians are conditioned at timestamp $t$, yielding 3D position and covariance. In contrast, FreeTimeGS (FTGS) allows Gaussian primitives to be placed freely in space and time, enabling flexible modeling of complex dynamic scenes. Each Gaussian is assigned an explicit motion function based on velocity and a temporal opacity function to control its position and visibility over time. Spacetime Gaussians (STG) models dynamic scenes using Gaussian primitives with polynomial motion and rotation functions, along with temporal opacity to control their visibility over time. For further details, please refer to the original paper. 

Despite their different temporal modeling strategies, all three representations produce time-dependent 3D Gaussian primitives for rendering. This shared Gaussian-based structure enables our framework to operate on individual primitives independently of their temporal parameterizations. We denote the set of Gaussian primitives in a pretrained dynamic scene representation as $\mathcal{P}={(x_i,f_i,a_i)}_{i=1}^{N}$, where $x_i$ denotes the reference spatial position, $f_i$ represents the appearance features, and $a_i$ contains the remaining per-primitive attributes, including opacity, scale, rotation, and temporal parameters.

\section{Methods}
\begin{figure}[t]
\begin{center}
\includegraphics[width=1.0\linewidth]{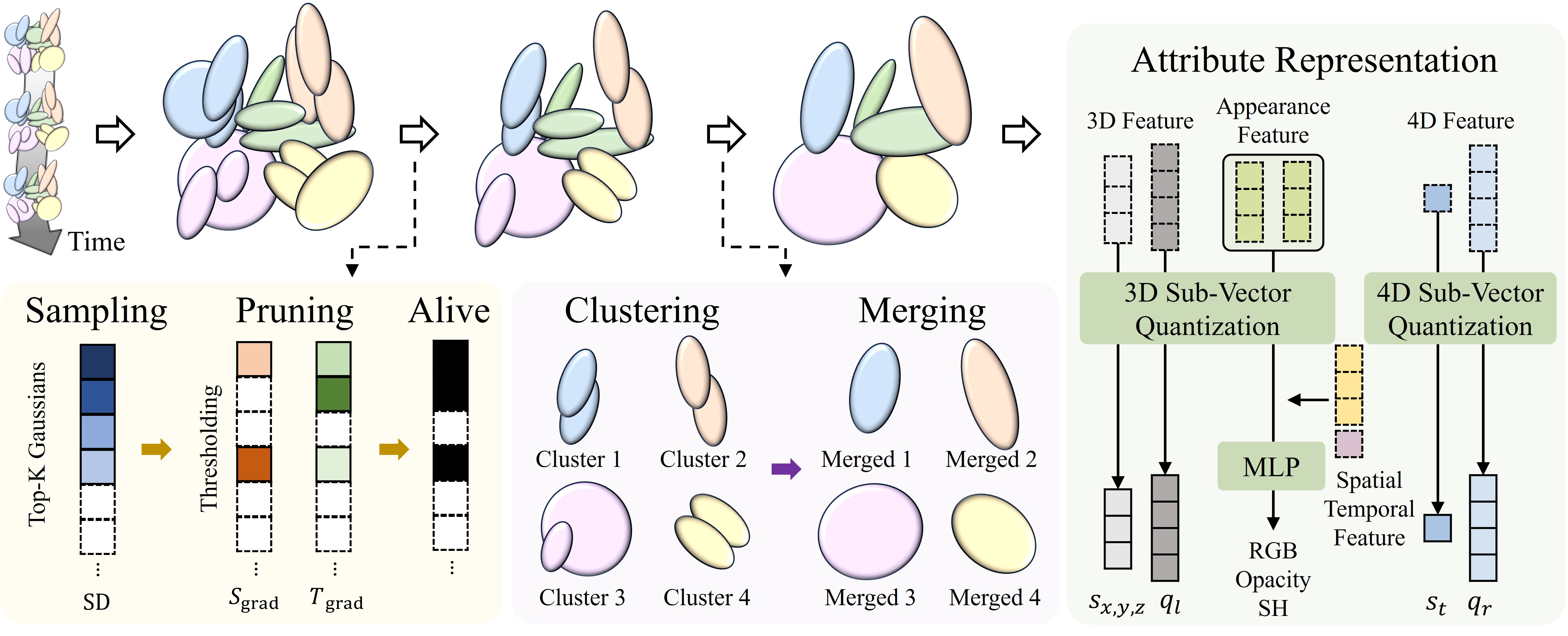}
\end{center}
\vspace{-0.2cm}
   \caption{The overall architecture of the proposed \textit{OMG4}.}
\vspace{-0.2cm}
\label{fig:main}
\end{figure}

We propose a novel framework for high-fidelity dynamic scene representation with a compact set of Gaussians consisting of Gaussian Sampling (\cref{sec:Gaussian_Sampling}), Gaussian Pruning (\cref{sec:Gaussian_Pruning}), Gaussian Merging (\cref{sec:Gaussian_Merging}), and Gaussian attributes compression (\cref{sec:att_comp}). An overview of the entire pipeline is provided in \cref{fig:main}.

\begin{figure}[t]
\begin{center}
\includegraphics[width=1.0\linewidth]{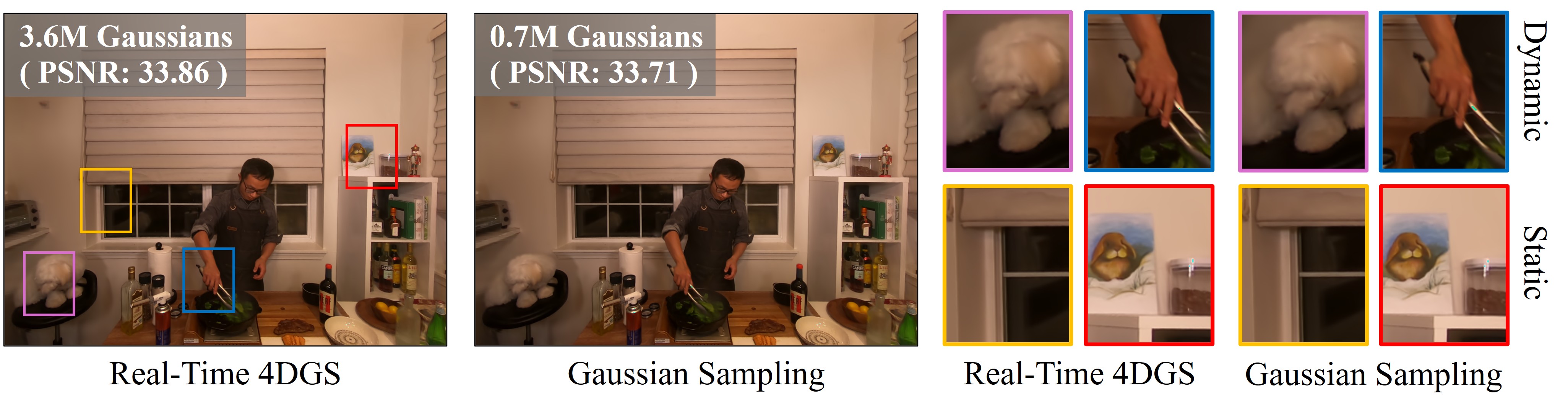}
\end{center}
\vspace{-0.2cm}
   \caption{A comparison of rendered images. (Left) Real-Time4DGS~\cite{realtime-4dgs} with 3.6M Gaussians. (Right) 1K optimization after Gaussian Sampling with 0.7M Gaussians. }
\vspace{-0.2cm}
\label{fig:gaussian_sampling}
\end{figure}

\subsection{Gaussian Sampling}
\label{sec:Gaussian_Sampling}
Primitive redundancy in dynamic scene representation is heterogeneous across space and time. Although recent studies observed that many Gaussians contribute only marginally to reconstruction quality~\cite{OMG,lp3dgs}, such redundancy becomes more structured in dynamic scenes due to the distinct spatio-temporal characteristics of static and dynamic regions.
Static regions are typically represented by temporally persistent and spatially dispersed Gaussians, while dynamic regions contain temporally localized and spatially dense Gaussians. 
Thus, spatial-only importance criteria can either over-preserve static background primitives or remove primitives essential for dynamic motion. 
We therefore propose the Static--Dynamic Score (SD-Score), which evaluates each Gaussian using both spatial rendering sensitivity and temporal consistency. 
As shown in~\cref{fig:gaussian_sampling}, SD-Score selects a compact set of salient primitives from a pretrained Real-Time4DGS model~\cite{realtime-4dgs} while preserving reconstruction quality.

The proposed Static–Dynamic Score (SD-Score) to calculate the importance of each Gaussian is defined as follows:
\begin{equation}
\label{eq:sd_score}
\textrm{SD}^{(i)} = \mathcal{N}_{\mathrm{minmax}}\!\left(S_{\mathrm{grad}}^{(i)}\right) \cdot \mathcal{N}_{\mathrm{rank}}\!\left(T_{\mathrm{grad}}^{(i)}\right), \quad S_{\textrm{grad}}^{(i)} = \sum_{j=1}^{N} \vert\vert \nabla_{u_{i,j}} L_j \vert\vert_2, \quad T_{\textrm{grad}}^{(i)} = \sum_{j=1}^{N} \nabla_{\mu_{t_i}} L_j.
\end{equation}



Here, $N$ is the number of training images (\# of input views $\times$ \# of frames), $L_j$ is the reconstruction loss for image $j$, $u_{i,j}\in\mathbb{R}^2$ denotes the projected 2D coordinate of Gaussian $i$ in image $j$, and $\mu_{t_i}\in\mathbb{R}$ is its temporal coordinate. $\mathcal{N}_{\mathrm{minmax}}(\cdot)$ and $\mathcal{N}_{\mathrm{rank}}(\cdot)$ denote min-max normalization and rank normalization, respectively.
Since signed temporal gradients are sensitive to outliers, we apply rank normalization based on their relative ordering, while normalizing both scores to comparable scales before multiplication.
\vspace{-0.3cm}

\paragraph{Static Score.}
The static score $S_{\textrm{grad}}^{(i)}$ accumulates screen-space gradients over all training images, measuring the spatial rendering sensitivity of each Gaussian with respect to its projected coordinates. In static regions, Gaussians are temporally long-lived yet relatively sparse, so even small positional perturbations can consistently affect the reconstruction loss across many frames, resulting in high static scores. In contrast, Gaussians in dynamic regions are often temporally localized and spatially dense. Consequently, the effect of a single Gaussian’s positional change on the overall reconstruction loss is diluted, resulting in lower static scores.
\vspace{-0.3cm}
\paragraph{Dynamic Score.}
The dynamic score $T_{\textrm{grad}}^{(i)}$ is computed by summing the gradients with respect to the temporal coordinate of each Gaussian, capturing a signed temporal trend in the reconstruction loss. Gaussians with high normalized dynamic scores tend to be important for dynamic regions, where small temporal shifts can substantially affect motion and transient appearance. 
We accumulate signed time gradients rather than their magnitudes, which acts as an implicit consistency filter: temporally inconsistent Gaussians with frequent gradient sign flips are suppressed through cancellation, whereas Gaussians with coherent temporal trends are better preserved after normalization.

By combining the complementary static and dynamic scores, the SD-Score provides a spatio-temporal importance measure tailored to explicit 4DGS approaches. We sample the Gaussians with high $\mathrm{SD}$ values using a sampling ratio of $\tau_{\textrm{GS}}$, forming the sampled set $\mathcal{P}_{\textrm{GS}}$, and optimize it for $T_{\textrm{GS}}$ iterations.

\begin{figure}[t]
\begin{center}
\includegraphics[width=1.0\linewidth]{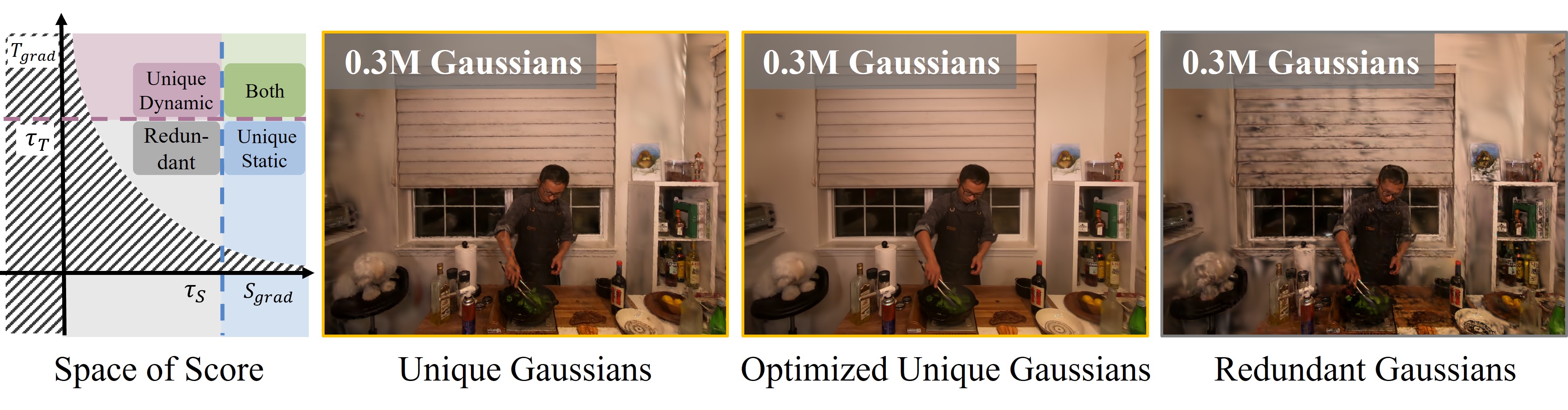}
\end{center}
\vspace{-0.6cm}
   \caption{ Illustration of Gaussian Pruning. (Left) Space defined by Static and Dynamic score and Gaussian Sampling Boundary. (Middle) A rendered image with unique Gaussians that satisfy both unique static and dynamic thresholds, and one after 1K optimization. (Right) A rendered image with redundant Gaussians that are not included in a unique area. }
\vspace{-0.2cm}
\label{fig:gaussian_pruning}
\end{figure}

\subsection{Gaussian Pruning}
\label{sec:Gaussian_Pruning}
Although Gaussian Sampling selects primitives with high SD-Scores, the sampled set $\mathcal{P}_{GS}$ can still contain redundant Gaussians. In response, we introduce a Gaussian Pruning strategy that further refines the set by eliminating superfluous Gaussians that remain after the first stage.

\cref{fig:gaussian_pruning} illustrates the space defined by $\mathcal{N}_{\mathrm{minmax}}(S_{\textrm{grad}})$ and $\mathcal{N}_{\mathrm{rank}}(T_{\textrm{grad}})$, where the curve $\mathcal{N}_{\mathrm{rank}}(T_{\textrm{grad}}) = c / \mathcal{N}_{\mathrm{minmax}}(S_{\textrm{grad}})$ represents the selection boundary from Gaussian Sampling, with $c$ denoting the SD-Score threshold determined by the sampling ratio $\tau_{\textrm{GS}}$. Within this space, we identify Gaussians that are likely redundant, and our empirical analysis shows that those with both $S_{\textrm{grad}}^{(i)}$ and $T_{\textrm{grad}}^{(i)}$ values being small can be safely filtered out (gray regions in \cref{fig:gaussian_pruning}). To this end, we apply thresholding to both scores, with each threshold defined as follows:

\begin{equation}
\begin{aligned}
\mathcal{P}_{GP}
&= \left\{ G_i \in \mathcal{P}_{GS} \,\middle|\,
\left(\mathcal{N}(S_{\text{grad}}^{(i)}) \ge \tau_S\right) \lor
\left(\mathcal{N}(T_{\text{grad}}^{(i)}) \ge \tau_T\right)
\right\},\\
\tau_S &= Q_p\!\left(\left\{ \mathcal{N}(S_{\text{grad}}^{(i)}) \right\}_{G_i\in\mathcal{P}_{GS}}\right), \qquad
\tau_T = Q_p\!\left(\left\{ \mathcal{N}(T_{\text{grad}}^{(i)}) \right\}_{G_i\in\mathcal{P}_{GS}}\right),
\end{aligned}
\end{equation}

where $\tau_S$ and $\tau_T$ are the thresholds for $S_{\textrm{grad}}$ and $T_{\textrm{grad}}$ respectively, and $Q_p(\cdot)$ denotes the $p$-quantile value computed over the sampled set $\mathcal{P}_{GS}$. As shown in \cref{fig:gaussian_pruning}, the pruned set, $\mathcal{P}_{GP}$, can still represent the scene with fine details and we optimize $\mathcal{P}_{GP}$ for $T_{GP}$ iterations.

One may question the necessity of employing two distinct stages, since, in principle, sampling and pruning could be integrated into a single process. Nevertheless, our experiments indicate that separating these stages proves to be more effective, as evidenced by the ablation study in \cref{sec:abl}. 

\subsection{Gaussian Merging}
\label{sec:Gaussian_Merging}
While Gaussian Sampling and Gaussian Pruning are effective in preserving a compact and salient set of Gaussians, both stages evaluate the importance of each Gaussian individually without considering similarities across Gaussians. To further reduce redundancy, we introduce a Gaussian Merging technique that clusters highly similar Gaussians and fuses them into single representative Gaussians.

We first partition the 4D space-time volume into grid cells and perform clustering only among Gaussians within the same cell. This local grouping reduces the computational cost of pairwise comparison and prevents merging Gaussians that are spatially or temporally mismatched. Within each grid cell, we define the similarity score $S(G_i, G_j)$ between two Gaussians as a sum of spatial proximity and appearance similarity:
\begin{equation}
S(G_i,G_j) = -(||{x}_i-{x}_j||^2_2 + \lambda||{f}_i-{f}_j||^2_2),
\end{equation}
where $\lambda$ is a fixed balancing weight, ${x}_i \in \mathbb{R}^3$ is the position, and ${f}_i \in \mathbb{R}^3$ is the zero-th order spherical-harmonics (RGB) coefficient of $G_i$. A higher score indicates a greater spatial-appearance similarity. 
For each Gaussian $G_i$, we construct a cluster of Gaussians $C_i$ by thresholding the similarity score with $\tau_{\textrm{sim}}$:
\begin{equation}
C_i = \{ i \} \cup \{ j \in \mathcal{P}_i \mid S(G_i,G_j) \ge \tau_{\textrm{sim}} \}, 
\end{equation}
where $\mathcal{P}_i$ denotes a set of Gaussian indices within the spatio-temporal grid cell that contain the Gaussian $G_i$. We then deduplicate identical clusters and remove clusters that are subsets of larger ones, yielding a final set of maximal clusters $\mathcal{C} = \{C_q\}_{q=1}^{N_C}$. 
Gaussians that do not belong to any non-singleton cluster are kept unchanged.

Given the resulting clusters, we merge the Gaussians within each cluster into a single representative Gaussian using learnable weights for position and appearance.
For each cluster $C_q$, we introduce learnable per-Gaussian weights $w_i^x$ and $w_i^f$ for position and appearance, respectively. 
During training-time rendering, all Gaussians in $C_q$ are replaced by a single proxy Gaussian:
\begin{equation}
\bar{x}_q = \sum\limits_{i\in C_q} {w}_{i}^{x} {x}_i, \quad \bar{f}_q = \sum\limits_{i\in C_q} {w}_{i}^{f} {f}_i, \quad \bar{a}_q = a_{r(C_q)}, \quad r(C_q)= \arg\max_{i \in C_q} {w}_{i}^{x},
\end{equation}
where both weights $w_i^x, w_i^f$ are normalized per cluster. The proxy position and appearance are computed as weighted sums of the Gaussians within the cluster, while the remaining attributes $\bar{a}_q$ are taken from the representative Gaussian with the largest position weight. We use the attributes of a representative Gaussian instead of averaging all attributes, because parameters such as scale, rotation, and temporal attributes affect rendering nonlinearly; naively averaging them can yield unstable or poorly representative Gaussians. We optimize both weights $w_i^x, w_i^f$ for $T_{GM}$ iterations and repeat Gaussian Merging $M$ times while progressively increasing the grid size, enabling the model to merge local redundancies first and then consider broader spatio-temporal neighborhoods. The detailed process of Gaussian Merging is provided in the supplementary material.

\subsection{Attribute Compression}
\label{sec:att_comp}
Given the compact Gaussian set, we compress attributes by adapting the OMG architecture~\cite{OMG}, which is initially designed for static 3DGS, to the 4DGS setting with explicit time conditioning. 
OMG employs an MLP to extract spatial features and uses three lightweight MLPs to predict opacity, static color, and view-dependent color from the spatial features and position.  To handle dynamic scenes, we additionally condition these MLPs on the time coordinate, enabling time-varying opacity and appearance. This compression stage is applied after Gaussian reduction because directly evaluating the MLP-based implicit attribute prediction over the full set of millions of 4D Gaussians introduces substantial memory overhead. 
Thus, sufficient Gaussian reduction is a necessary prerequisite for tractable attribute compression in explicit 4DGS.

We further extend Sub-Vector Quantization (SVQ)~\cite{OMG}, which partitions an input vector into sub-vectors and quantizes each with a small codebook, to compress the attributes of 4D Gaussians. We retain the 4D Gaussian means in full precision for stability, and apply SVQ to the remaining attributes, including the additional rotation quaternion and temporal-axis scales introduced in Real-Time4DGS~\cite{realtime-4dgs} to model scene motion. However, quantizing both static and dynamic attributes simultaneously leads to unstable optimization, since perturbations to temporal attributes can affect the time-varying geometry and appearance of dynamic primitives, thereby interfering with spatial reconstruction and temporal consistency. To mitigate this, we propose a staged SVQ scheme. We first apply SVQ only to 3D attributes and optimize the compressed representation for $T_{3D}$ iterations, allowing spatial geometry and appearance to stabilize. 
We then activate SVQ for 4D attributes, enabling temporal attributes to be compressed on top of the stabilized spatial representation. This staged scheme mitigates instability from simultaneous spatial-temporal quantization. Finally, we compress the quantized elements using Huffman encoding~\cite{huffman} followed by LZMA compression~\cite{lzma}.

\begin{table*}[t!]
    \centering
    \caption{Quantitative results on N3DV~\cite{n3dv}  at $1352 \times 1014$ resolution. 
    Baseline results are sourced from the corresponding papers unless marked with $^*$; 
    $^*$ denotes reproduced results. 
    $^\bullet$ denotes results excluding the \textit{Coffee Martini} scene.}
    \resizebox{\linewidth}{!}{
    \begin{tabular}{lcccccc}
    \toprule
    Method & PSNR$\uparrow$ & SSIM$\uparrow$ & LPIPS (A)$\downarrow$ & FPS$\uparrow$ & \# G$\downarrow$ & Storage (MB)$\downarrow$ \\
    \midrule
    K-Planes~\cite{K-Planes} & 31.63 & -- & -- & 0.3 & -- & 311 \\
    HexPlane~\cite{Hexplane}$^\bullet$ & 31.71 & -- & 0.075 & 0.2 & -- & 250 \\
    \midrule
    Real-Time4DGS~\cite{realtime-4dgs}$^*$ & 31.96 & 0.946 & 0.051 & 57 & 3,397,510 & 2087 \\
    STG~\cite{spacetimegaussian} & 32.05 & 0.946 & 0.044 & 140 & -- & 200 \\
    4DGS~\cite{4dgs} & 31.15 & -- & 0.049 & 82 & -- & 18.0 \\
    MEGA~\cite{MEGA} & 31.49 & -- & 0.057 & 77 & -- & 25.05 \\
    CSTG~\cite{CSTG} & 31.69 & 0.945 & 0.054 & 186 & -- & 15.4 \\
    GIFStream~\cite{GIFStream}--L & 31.75 & 0.938 & 0.051 & 95 & -- & 10.0 \\
    GIFStream~\cite{GIFStream}$^*$--S & 31.25 & 0.937 & 0.057 & 156 & -- & 6.63 \\
    4DGS-CC (4DGS)~\cite{4DGS-CC}--L & 31.54 & -- & 0.062 & -- & -- & 3.20 \\ 
    4DGS-CC (4DGS)~\cite{4DGS-CC}--S & 30.97 & -- & 0.065 & -- & -- & 2.30 \\ 
    4DGS-CC (Saro-GS)~\cite{4DGS-CC}--L & 31.93 & -- & 0.045 & -- & -- & 20.9 \\ 
    4DGS-CC (Saro-GS)~\cite{4DGS-CC}--S & 31.60 & -- & 0.047 & -- & -- & 17.0 \\ 
    ADC-GS~\cite{ADC-GS}$^\bullet$--L & 31.67 & -- & 0.061 & 110 & -- & 6.57 \\
    ADC-GS~\cite{ADC-GS}$^\bullet$--S & 31.41 & -- & 0.066 & 126 & -- & 4.04 \\
    \midrule
    Ours--L & 32.03 & 0.942 & 0.058 & 118 & 228,494 & 4.71 \\
    Ours--M & 31.85 & 0.941 & 0.060 & 129 & \cellcolor[HTML]{FFFC9E}170,044 & \cellcolor[HTML]{FFFC9E}3.59 \\
    Ours--S & 31.75 & 0.939 & 0.062 & 133 & \cellcolor[HTML]{FFCCC9}135,684 & \cellcolor[HTML]{FFCCC9}2.93 \\
    Ours--T & 31.54 & 0.936 & 0.067 & 142 & \cellcolor[HTML]{FFCE93}97,186 & \cellcolor[HTML]{FFCE93}2.18 \\
    Ours$^\bullet$--L & 32.76 & 0.949 & 0.052 & 123 & 214,106 & 4.43 \\
    Ours$^\bullet$--T & 32.25 & 0.943 & 0.063 & 146 & 90,930 & 2.05 \\
    \bottomrule
    \end{tabular}
    }
    \label{tab:quant_n3dv}
    \vspace{-0.25cm}
\end{table*}

\section{Experiments and Results}
\subsection{Implementation Details}
For the reduction stages, we optimize each stage for $1{,}000$ iterations ($T_{GS}=T_{GP}=T_{GM}=1{,}000$): Gaussian Sampling is followed by Gaussian Pruning at the $1{,}000^\text{th}$ iteration, and Gaussian Merging is applied at the $2{,}000^\text{th}$ and $3{,}000^\text{th}$ iterations. 
Attribute compression starts at the $4{,}000^\text{th}$ iteration, with MLP training first, 3D SVQ activated at the $9{,}000^\text{th}$ iteration, and 4D SVQ at the $10{,}000^\text{th}$ iteration. 
For the medium-size model (\textit{Ours--M}), we set $\tau_{GS}=0.2$, use a 0.8-quantile pruning threshold, and perform Gaussian Merging twice with a spatial grid scaling factor of 1.2 and temporal grid size of 2.0. 
We use identical hyperparameters for all scenes within each dataset. Reported storage includes all rendering components, including Gaussian attributes, MLP weights, codebooks, and entropy-coding metadata. More details are provided in the supplementary material.
\vspace{-0.2cm}

\begin{table}[t]
    \centering
    \caption{Quantitative results of \textit{OMG4} on FTGS (FreeTimeGS)~\cite{freetimegs}, evaluated on N3DV~\cite{n3dv}.}
    \vspace{-0.2cm}
    \small
    \setlength{\tabcolsep}{3pt}
    \begin{tabular}{lcccccc}
    \toprule
    Method & PSNR$\uparrow$ & SSIM$\uparrow$ & LPIPS (A)$\downarrow$ & FPS$\uparrow$ & \# G$\downarrow$ & Storage (MB)$\downarrow$ \\
    \midrule
    FTGS & 32.80 & 0.958 & 0.040 & 532 & 500,000 & 61.04 \\
    \midrule
    Ours--L & 32.71 & 0.955 & 0.044 & 346 & 146,891 & 4.06 \\
    Ours--M & 32.62 & 0.954 & 0.046 & 354 & 109,191 & 3.07 \\
    Ours--S & 32.45 & 0.951 & 0.050 & 374 & \cellcolor[HTML]{FFFC9E}80,673 & \cellcolor[HTML]{FFFC9E}2.32 \\
    Ours--T & 32.28 & 0.950 & 0.051 & 379 & \cellcolor[HTML]{FFCE93}65,706 & \cellcolor[HTML]{FFCE93}1.93 \\
    \bottomrule
    \end{tabular}
    \label{tab:ftgs}
    \vspace{-0.2cm}
\end{table}

\begin{table}[t]
    \centering
    \caption{Quantitative results on SelfCap~\cite{selfcap}.}
    \vspace{-0.2cm}
    \small
    \setlength{\tabcolsep}{3pt}
    \begin{tabular}{lcccccc}
    \toprule
    Method & PSNR$\uparrow$ & SSIM$\uparrow$ & LPIPS (V)$\downarrow$ & FPS$\uparrow$ & \# G$\downarrow$ & Storage (MB)$\downarrow$ \\
    \midrule
    FTGS~\cite{freetimegs} & 26.96 & 0.908 & 0.194 & 241 & 1,000,000 & 122 \\
    GIFStream~\cite{GIFStream}$^*$--L & 25.83 & 0.876 & 0.262 & 94 & -- & 7.09 \\
    GIFStream~\cite{GIFStream}$^*$--S & 25.56 & 0.866 & 0.284 & 104 & -- & \cellcolor[HTML]{FFCCC9}5.52 \\
    \midrule
    Ours (FTGS)--L & 26.93 & 0.897 & 0.224 & 218 & 293,402 & 7.50 \\
    Ours (FTGS)--M & 26.91 & 0.897 & 0.228 & 235 & \cellcolor[HTML]{FFFC9E}261,630 & 6.72 \\
    Ours (FTGS)--S & 26.88 & 0.896 & 0.238 & 265 & \cellcolor[HTML]{FFCCC9}201,983 & \cellcolor[HTML]{FFFC9E}5.24 \\
    Ours (FTGS)--T & 26.74 & 0.892 & 0.254 & 285 & \cellcolor[HTML]{FFCE93}154,155 & \cellcolor[HTML]{FFCE93}4.06 \\
    \bottomrule
    \end{tabular}
    \label{tab:quant_selfcap}
    \vspace{-0.4cm}
\end{table}

\begin{table*}[t]
    \centering
    \caption{Quantitative results on ENeRF-Outdoor~\cite{enerf}. 
    $N$ denotes the number of models, and $F$ denotes the number of frames in each chunk.}
    \vspace{+0.1cm}
    \scriptsize
    \setlength{\tabcolsep}{2.5pt}
    \resizebox{\textwidth}{!}{
    \begin{tabular}{@{}llcccccc@{}}
    \toprule
    Setting & Method & PSNR$\uparrow$ & SSIM$\uparrow$ & LPIPS (V)$\downarrow$ & FPS$\uparrow$ & \#G$\downarrow$ & Storage (MB)$\downarrow$ \\
    \midrule
    \multirow{3}{*}{$N=1, F=1200$}
    & Real-Time4DGS~\cite{realtime-4dgs} & \multicolumn{6}{c}{OOM} \\
    & FTGS~\cite{freetimegs} & 24.95 & 0.686 & 0.310 & 317 & 4,000,000 & 488 \\
    & Ours (FTGS) & 24.76 & 0.706 & 0.280 & 140 & 1,999,800 & 52.29 \\
    \midrule
    \multirow{5}{*}{$N=4, F=300$}
    & Real-Time4DGS~\cite{realtime-4dgs} & 24.40 & 0.689 & 0.293 & 66 & 13,121,660 & 8059 \\
    & FTGS~\cite{freetimegs} & 25.04 & 0.701 & 0.287 & 328 & 4,000,000 & 488 \\
    & GIFStream~\cite{GIFStream} & 23.20 & 0.662 & 0.298 & 44 & -- & 85.85 \\
    & Ours (RT4DGS) & 24.30 & 0.660 & 0.351 & 111 & \cellcolor[HTML]{FFCE93}515,443 & \cellcolor[HTML]{FFCE93}11 \\
    & Ours (FTGS) & 24.70 & 0.677 & 0.335 & 181 & \cellcolor[HTML]{FFCCC9}1,999,800 & \cellcolor[HTML]{FFCCC9}53.44 \\
    \midrule
    $N=20, F=60$
    & GIFStream~\cite{GIFStream} & 24.23 & 0.682 & 0.280 & 52 & -- & 175 \\
    \bottomrule
    \end{tabular}
    }
    \label{tab:quant_enerf}
    \vspace{-0.35cm}
\end{table*}

\begin{figure}[t]
\begin{center}
\includegraphics[width=1.0\linewidth]{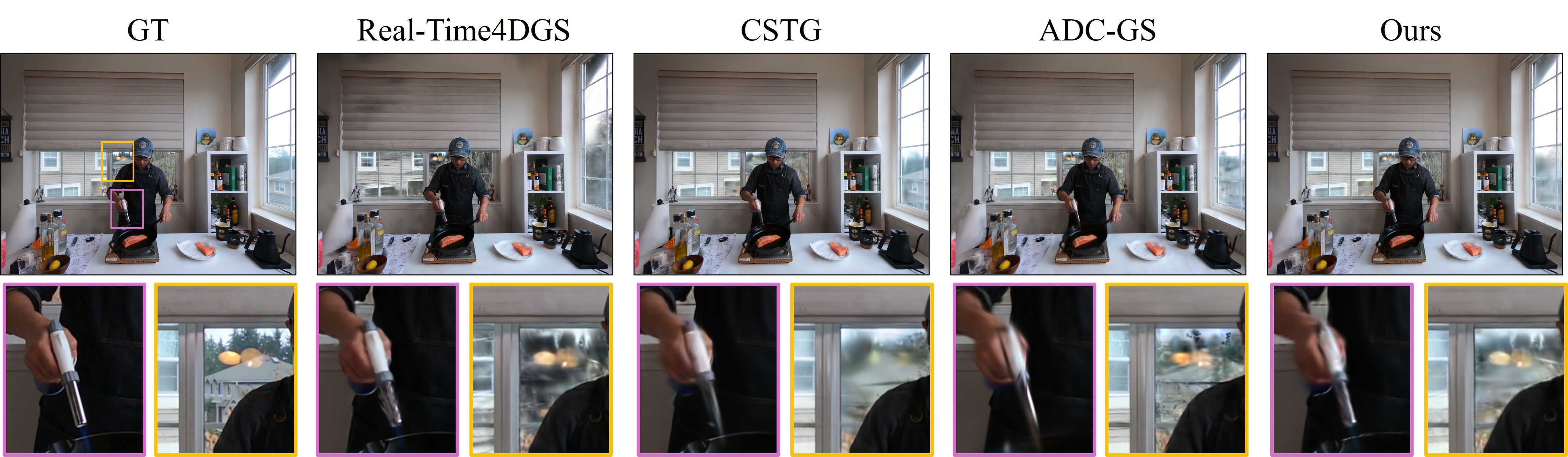}
\end{center}
\vspace{-0.2cm}
   \caption{Qualitative results on N3DV dataset~\cite{n3dv}.}
\vspace{-0.4cm}
\label{fig:quali_n3dv}
\end{figure}

\subsection{Results and Analysis}
We first conduct experiments on N3DV~\cite{n3dv}. As shown in~\cref{tab:quant_n3dv} and~\cref{fig:quali_n3dv}, we effectively reduce the model size of the baseline Real-Time4DGS~\cite{realtime-4dgs}, from 2 GB to around 3 MB, while preserving comparable visual quality (\textit{Ours--M}), corresponding to a 99\% reduction in storage. Compared to a recent state-of-the-art method GIFStream~\cite{GIFStream}, \textit{OMG4} achieves 65\% reduction in storage (from 10 MB to 3.59 MB) while achieving comparable reconstruction fidelity 
and slightly higher PSNR (31.75 vs. 31.85). In the low-memory regime, \textit{Ours--T} achieves a better PSNR-storage trade-off than 4DGS-CC~\cite{4DGS-CC}, with higher PSNR (31.54 vs. 30.97) and smaller storage (2.18 MB vs. 2.30 MB).
Beyond compression, \textit{OMG4} improves FPS by 2.26$\times$ over the baseline due to its compact representation. 
We provide rate-distortion curves and BD-rate analyses in the supplementary material.

\subsubsection{Application on FreeTimeGS}
We further extend \textit{OMG4} to FreeTimeGS~\cite{freetimegs} to evaluate its generality beyond Real-Time4DGS~\cite{realtime-4dgs}. 
Starting from the pretrained FreeTimeGS models\footnote{We reproduced FreeTimeGS models since the codes and pre-trained models were not publicly available. All experiments were conducted with our reproduced models.} (FTGS), \textit{OMG4} reduces Gaussians from 500K to around 147K (Ours--L) and 66K (Ours--T) primitives. As shown in~\cref{tab:ftgs}, \textit{OMG4} reduces the storage of FreeTimeGS by up to 90\% while maintaining comparable reconstruction quality, and remains effective even under a strict memory budget of under 2 MB. 
This result highlights the versatility of \textit{OMG4}, showing that it can be effectively applied to diverse 4D representations. Due to the page limit, results on extending our method to STG~\cite{spacetimegaussian} are provided in the supplementary material.

\subsubsection{Results on Challenging Datasets}
Additionally, we conduct experiments under more challenging scenarios, including complex non-rigid motion in SelfCap~\cite{selfcap} and long-sequence reconstruction in ENeRF-Outdoor~\cite{enerf}. \cref{tab:quant_selfcap} summarizes results on the SelfCap~\cite{selfcap} dataset. It consists of six scenes, and we use 60 frames for evaluation following the FreeTimeGS~\cite{freetimegs} protocol.\footnote{We do not report Real-Time4DGS-based results because it does not provide a reliable pretrained representation under this protocol, making subsequent compression by \textit{OMG4} uninformative.} 
\textit{OMG4} reduces FTGS from 122 MB to 7.5 MB with a reasonable fidelity trade-off under aggressive compression. 
Compared to GIFStream~\cite{GIFStream}, \textit{Ours--T} achieves better reconstruction quality across all metrics with less storage than GIFStream, improving PSNR from 25.56 to 26.74 while reducing storage from 5.52 MB to 4.06 MB.
Please refer to the supplementary material for qualitative comparisons.

\cref{tab:quant_enerf} reports results on ENeRF-Outdoor~\cite{enerf} dataset (1200 frames), evaluating long-sequence scenarios. Since Real-Time4DGS~\cite{realtime-4dgs} and GIFStream~\cite{GIFStream} cannot be trained on all frames at once due to OOM, we report their results using temporal chunking with four 300-frame models. For a fair comparison, we also evaluate FreeTimeGS~\cite{freetimegs} and \textit{OMG4} under the same 300-frame chunking protocol. In addition, we follow the original 60-frame chunk setting of GIFStream for comparison. Despite the longer temporal span, \textit{OMG4} achieves visually satisfactory reconstruction quality with a compact memory footprint, demonstrating its robustness under long-sequence scenarios. Notably, the FreeTimeGS-based \textit{OMG4} can also be trained on the full 1200-frame sequence in a single run within an hour, whereas GIFStream requires training 20 chunk-specific models, taking 17 hours in total. This highlights the practical scalability of our method for long dynamic sequences.

\subsubsection{Ablation Studies}
\label{sec:abl}
We present ablation studies on each module of \textit{OMG4} in \cref{tab:abl_method}. Directly applying Attribute Compression to the unreduced Real-Time4DGS~\cite{realtime-4dgs} results in out-of-memory failure due to the MLP overhead over millions of primitives, showing that Gaussian reduction is necessary before attribute compression.
When applying Gaussian Sampling before attribute compression, we remove 80\% of the Gaussians while achieving reconstruction quality comparable to Real-Time4DGS. This may be attributed to a regularization effect, where removing noisy or redundant primitives helps the remaining Gaussians converge more cleanly~\cite{ali2024trimmingfatefficientcompression}. 
Adding Gaussian Pruning further reduces storage to a few megabytes, and Gaussian Merging achieves the minimal number of Gaussians. We also ablate the SVQ schedule inside Attribute Compression. Compared with simultaneous 3D/4D quantization, the proposed staged 3D-to-4D SVQ improves PSNR by 0.31 dB under a similar storage budget, supporting our design of stabilizing spatial attributes before compressing temporal attributes. It may appear possible to perform Gaussian Sampling and Gaussian Pruning simultaneously. However, this variant results in a PSNR drop of 0.15 dB (\textit{GS}+\textit{GP}). This highlights the importance of the proposed multi-stage pipeline: the intermediate optimization after Gaussian Sampling helps stabilize the reduced representation before further pruning. Therefore, our pipeline separates Gaussian Sampling and Gaussian Pruning with an intermediate optimization period. Additional ablation studies are provided in the supplementary material.

\begin{table}[t]
\centering
\caption{Ablation study on each module and SVQ schedule. 
\textit{GS}, \textit{GP}, \textit{GM}, and \textit{AC} denote Gaussian Sampling, Gaussian Pruning, Gaussian Merging, and Attribute Compression, respectively. 
\textit{Simul.} denotes simultaneous 3D/4D SVQ, and \textit{Staged} denotes the proposed staged 3D-to-4D SVQ.}
\resizebox{\columnwidth}{!}{
\begin{tabular}{ccccclcccccc}
\toprule
\multicolumn{5}{c}{Modules} & SVQ & PSNR$\uparrow$ & SSIM$\uparrow$ & LPIPS$\downarrow$ & FPS$\uparrow$ & \# G$\downarrow$ & Storage (MB)$\downarrow$ \\
\cmidrule(lr){1-5} \cmidrule(lr){6-12}
GS+GP & GS & GP & GM & AC & Schedule & & & & & & \\
\midrule
 &  &  &  &  & -- & 31.96 & 0.946 & 0.051 & 57 & 3,397,510 & 2087 \\
 &  &  &  & \checkmark & -- & \multicolumn{6}{c}{OOM} \\
 & \checkmark &  &  & \checkmark & Staged & 32.02 & 0.945 & 0.053 & 82 & 679,502 & 13.25 \\
 & \checkmark & \checkmark &  & \checkmark & Staged & 31.94 & 0.943 & 0.056 & 144 & 241,278 & 4.83 \\
\checkmark &  &  & \checkmark & \checkmark & Staged & 31.70 & 0.941 & 0.060 & 121 & 170,149 & 3.59 \\ 
 & \checkmark & \checkmark & \checkmark & \checkmark & Simul. & 31.54 & 0.940 & 0.061 & 126 & 170,540 & 3.59 \\
 & \checkmark & \checkmark & \checkmark & \checkmark & Staged & 31.85 & 0.941 & 0.060 & 129 & 170,044 & 3.59 \\
\bottomrule
\end{tabular}
}
\label{tab:abl_method}
\vspace{-0.2cm}
\end{table}

\section{Conclusion}
We introduce \textit{OMG4}, a compact dynamic-scene representation that progressively reduces the number of primitives and compresses the Gaussian attributes.
\textit{OMG4} first constructs a compact yet expressive primitive set through Gaussian Sampling, Gaussian Pruning, and Gaussian Merging, which identify salient Gaussians, remove redundant primitives, and fuse correlated ones in space-time. 
On top of this reduced set, we incorporate implicit appearance encoding and extend Sub-Vector Quantization (SVQ) to 4D representations with a staged 3D-to-4D quantization scheme. 
Consequently, \textit{OMG4} compresses Real-Time4DGS by three orders of magnitude while maintaining comparable reconstruction fidelity. 
It also generalizes to other dynamic Gaussian representations, including FreeTimeGS and STG, achieving consistent compression gains with high reconstruction quality. 
These results show that jointly addressing spatio-temporal primitive redundancy and attribute compression is crucial for compact 4D scene representation, positioning \textit{OMG4} as a practical step toward compact and high-fidelity dynamic scene representations.

\section*{Acknowledgments}
This research was supported in part by the Ministry of Science and ICT (MSIT) of Korea, under the National Research Foundation (NRF) grant (RS-2024-00337548); the Institute of Information \& Communications Technology Planning \& Evaluation(IITP) grant funded by the Korea government (MSIT) (No.RS-2026-25517417, Development of Compression, Reconstruction, and Rendering Technologies for Free-viewpoint Media); the Institute of Information \& Communications Technology Planning \& Evaluation (IITP) grant funded by the Korea government (MSIT) (No. 2018-0-00207, Immersive Media Research Laboratory); the IITP grant funded by the Korea government (MSIT): Development of immersive video spatial computing technology for ultra-realistic metaverse services (No.2022-0-00022, RS-2022- II220022); the AI Seoul Tech Research Support Program of the Seoul Future Foundation.

\clearpage
\bibliographystyle{plainnat}
\bibliography{main}


\FloatBarrier
\clearpage
\appendix

\begin{center}
\Large{\bfseries Optimized Minimal 4D Gaussian Splatting for Efficient Dynamic Scene Representation: Supplementary Material}
\par\vspace{3ex}
\end{center}

\FloatBarrier

\section{Additional Experimental Results}
\subsection{Rate-Distortion Curves and BD-Rate Analysis}
\begin{figure}[h!]
\begin{center}
\includegraphics[width=1.0\linewidth]{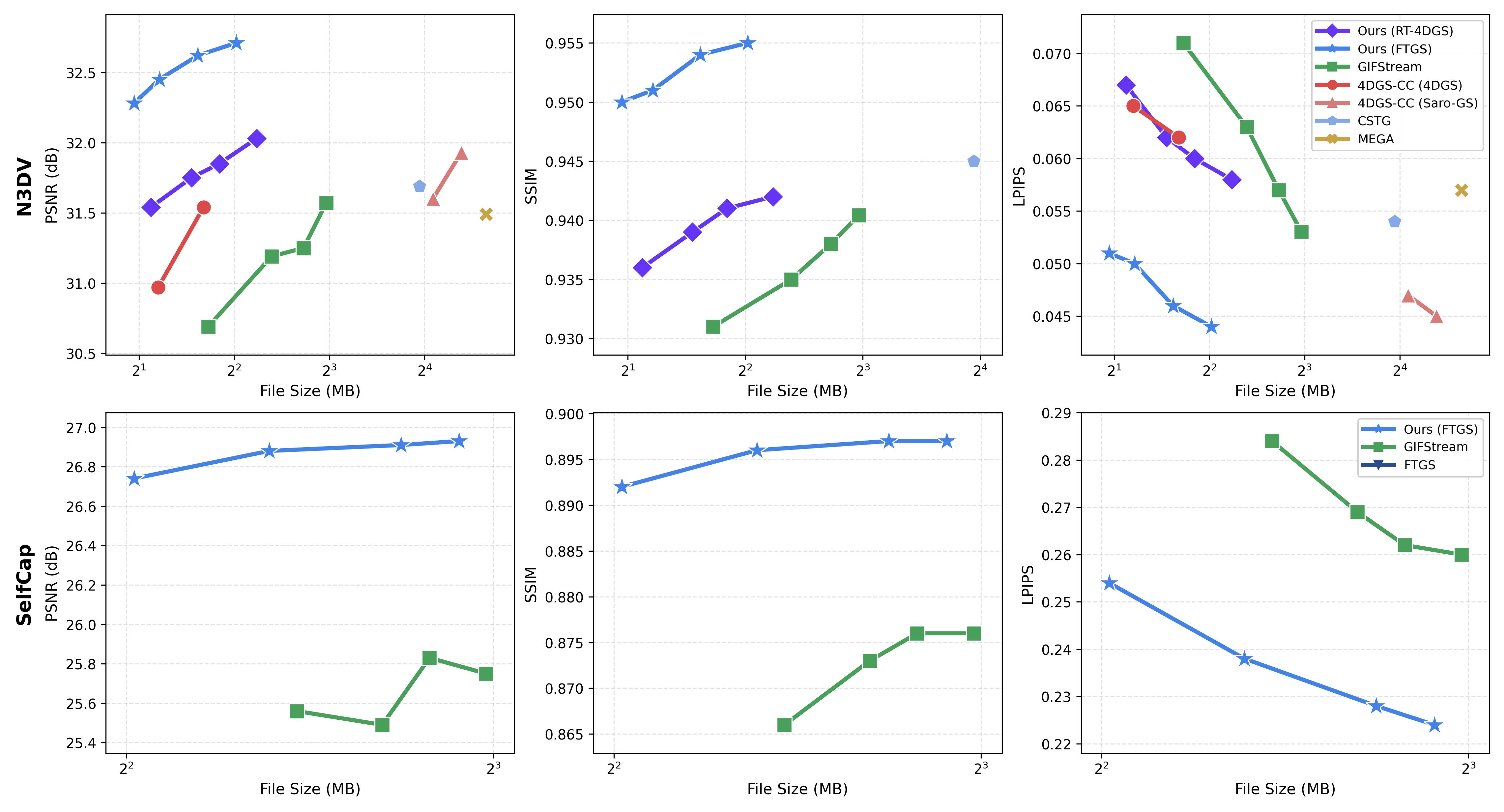}
\end{center}
\vspace{-0.5cm}
   \caption{Rate-distortion curves on N3DV~\cite{n3dv} (Top) and SelfCap~\cite{selfcap} (Bottom).}
\label{fig:rd_curves}
\end{figure}

\begin{table}[t]
\centering
\caption{PSNR-based BD-rate and BD-PSNR analysis. 
Storage size is used as the rate. 
Negative BD-rate indicates storage savings of \textit{OMG4} at the same PSNR, while positive BD-PSNR indicates higher PSNR at the same storage.}
\small
\setlength{\tabcolsep}{4pt}
\begin{tabular}{llcc}
\toprule
Dataset / Variant & Baseline & BD-Rate / PSNR (\%) $\downarrow$ & BD-PSNR (dB) $\uparrow$ \\
\midrule
N3DV / Ours (RT-4DGS) & GIFStream~\cite{GIFStream} & $-71.65$ & -- \\
N3DV / Ours (RT-4DGS) & 4DGS-CC (4DGS)~\cite{4DGS-CC} & -- & $+0.44$ \\
N3DV / Ours (RT-4DGS) & 4DGS-CC (Saro-GS)~\cite{4DGS-CC} & $-83.79$ & -- \\
N3DV / Ours (FTGS) & 4DGS-CC (4DGS)~\cite{4DGS-CC} & -- & $+1.30$ \\
SelfCap / Ours (FTGS) & GIFStream~\cite{GIFStream} & -- & $+1.30$ \\
\bottomrule
\end{tabular}
\label{tab:bd_rate}
\end{table}

We provide rate-distortion curves in~\cref{fig:rd_curves}, using storage size as the rate and PSNR, SSIM, and LPIPS as reconstruction quality metrics. 
On N3DV, the RT-4DGS-based \textit{OMG4} consistently achieves a favorable rate-distortion trade-off. 
Compared to GIFStream~\cite{GIFStream}, \textit{OMG4} attains higher PSNR and SSIM with substantially smaller storage. 
In the low-memory regime, it also provides a better PSNR-storage trade-off than 4DGS-CC (4DGS)~\cite{4DGS-CC}. 
The FTGS-based \textit{OMG4} further improves the overall RD curve, demonstrating that the proposed compression pipeline can benefit from stronger dynamic Gaussian backbones.

We compute PSNR-based BD-rate using the Bj{\o}ntegaard delta metric~\cite{bjontegaard2001calculation}, where storage size is used as the rate. 
BD-rate is computed over the overlapping PSNR range between each baseline and \textit{OMG4}; negative values indicate storage savings at the same PSNR. 
For the RT-4DGS-based \textit{OMG4}, the BD-rate results show substantial storage savings on N3DV: $-71.65\%$ compared to GIFStream, and $-83.79\%$ compared to 4DGS-CC (Saro-GS). 
When the PSNR ranges do not overlap but storage ranges overlap, we report BD-PSNR instead. 
For example, the FTGS-based \textit{OMG4} improves BD-PSNR by $+1.30$ dB over 4DGS-CC (4DGS) on N3DV.

On SelfCap, the FTGS-based \textit{OMG4} consistently outperforms GIFStream across the RD curves. 
Since the PSNR ranges of GIFStream and \textit{OMG4} do not overlap, PSNR-based BD-rate is not available. 
Instead, over the overlapping storage range, \textit{OMG4} improves BD-PSNR by $+1.30$ dB on average, further confirming its robustness on complex non-rigid motions.
\subsection{More Ablation Studies}
\begin{table}[t]
\centering
\caption{Ablation on the Dynamic Score of SD-Score.}
\small
\setlength{\tabcolsep}{5pt}
\begin{tabular}{lccc}
\toprule
Dynamic Score & PSNR$\uparrow$ & SSIM$\uparrow$ & LPIPS$\downarrow$ \\
\midrule
Absolute acc. & 31.65 & 0.941 & 0.060  \\
Signed acc. (Ours) & 31.85 & 0.941 & 0.060 \\
\bottomrule
\end{tabular}
\label{tab:abl_abs}
\end{table}
\paragraph{Ablation on Signed Temporal Gradients.}
\label{sec:abl_signed_tgrad}

We further analyze the design of the Dynamic Score in the proposed SD-Score. 
In the main method, the temporal component is computed by accumulating signed gradients with respect to the temporal coordinate:
\begin{equation}
T_{\mathrm{grad}}^{(i)} = \sum_{j=1}^{N} \nabla_{\mu_{t_i}} L_j.
\end{equation}
An alternative design is to accumulate gradient magnitudes:
\begin{equation}
T_{\mathrm{abs}}^{(i)} = \sum_{j=1}^{N} \left|\nabla_{\mu_{t_i}} L_j\right|.
\end{equation}
This variant measures the overall magnitude of temporal sensitivity, but ignores the directional consistency of temporal gradients across views and frames. 
As a result, Gaussians whose temporal gradients frequently change signs can still receive high scores, even though such sign changes may indicate unstable or flickering temporal behavior. In contrast, signed accumulation acts as an implicit temporal consistency filter. 
If a Gaussian exhibits inconsistent temporal gradients across views or frames, the positive and negative gradients are partially canceled before normalization, reducing its Dynamic Score. 
On the other hand, Gaussians with coherent temporal trends are better preserved. 
This behavior is desirable for aggressive compression, where the selected primitives should not only be sensitive to temporal changes but also provide stable temporal contributions. As shown in \cref{tab:abl_abs}, replacing the proposed signed temporal accumulation with absolute-gradient accumulation (\textit{Absolute acc.}) decreases PSNR from 31.85 to 31.65. This result supports our design choice: the signed temporal gradient provides a more effective ranking signal than magnitude-only temporal sensitivity for selecting dynamic primitives.

\begin{table}[t]
\centering
\caption{Ablation on Gaussian Merging design. 
The proposed method learns merging weights only for position and appearance and uses position-appearance similarity for clustering.}
\small
\setlength{\tabcolsep}{4pt}
\begin{tabular}{lccc}
\toprule
Variant & PSNR$\uparrow$ & SSIM$\uparrow$ & LPIPS$\downarrow$ \\
\midrule
Ours & 31.85 & 0.941 & 0.060 \\
+ Covariance-weighted merging & 31.72 & 0.941 & 0.060 \\
+ Temporal-weighted merging & 31.75 & 0.941 & 0.060 \\
\bottomrule
\end{tabular}
\label{tab:abl_merging_design}
\end{table}

\paragraph{Ablation on Gaussian Merging Design.}
\label{sec:abl_gaussian_merging}
We further analyze the design choice in Gaussian Merging. 
In the main method, we learn merging weights only for position and appearance, while the remaining attributes, such as scale, rotation, and temporal attributes, are inherited from the representative Gaussian. 
This design is motivated by the fact that position and RGB-like appearance are naturally defined in Euclidean spaces, making weighted combinations stable and interpretable. 
In contrast, scale, rotation, and temporal attributes affect rendering in a nonlinear manner, and naively blending them can produce poorly representative proxy Gaussians. To validate this design, we compare two variants. 
First, we additionally learn weights for covariance-related parameters, including scale and rotation. 
Second, we further learn weights for temporal attributes. 
As shown in~\cref{tab:abl_merging_design}, both variants slightly degrade reconstruction quality compared to the proposed design. 
This suggests that directly blending covariance-related or temporal attributes does not necessarily produce a better representative Gaussian, even though it increases the flexibility of the merging operation. 
Therefore, we use learnable weighted sums only for position and appearance, and inherit the remaining attributes from the representative Gaussian.

\paragraph{Hyperparameter Analysis for Gaussian Reduction.} 
\begin{table}[t]
    \centering
    \caption{Ablation study on the quantile level $p$ of Gaussian Pruning.}
    \small
    \setlength{\tabcolsep}{5pt}
    \begin{tabular}{lccccc}
    \toprule
    Setting & PSNR$\uparrow$ & SSIM$\uparrow$ & LPIPS$\downarrow$ & \# G$\downarrow$ & Storage (MB)$\downarrow$ \\
    \midrule
    $p = 0.6$ & 33.56 & 0.956 & 0.046 & 231,984 & 4.78 \\
    $p = 0.7$ & 33.55 & 0.956 & 0.046 & 192,789 & 4.02 \\
    $p = 0.8$ & 33.45 & 0.955 & 0.048 & 143,880 & 3.09 \\
    $p = 0.9$ & 32.94 & 0.949 & 0.057 & 82,605 & 1.89 \\
    \bottomrule
    \end{tabular}
    \label{tab:abl_GP}
\end{table}
\begin{table}[t]
    \centering
    \caption{Ablation study on the number of Gaussian Merging rounds $M$.}
    \small
    \setlength{\tabcolsep}{5pt}
    \begin{tabular}{lccccc}
    \toprule
    Setting & PSNR$\uparrow$ & SSIM$\uparrow$ & LPIPS$\downarrow$ & \# G$\downarrow$ & Storage (MB)$\downarrow$ \\
    \midrule
    $M = 1$ & 33.49 & 0.955 & 0.047 & 164,788 & 3.49 \\
    $M = 2$ & 33.45 & 0.955 & 0.048 & 143,880 & 3.09 \\
    $M = 3$ & 33.43 & 0.954 & 0.049 & 134,637 & 2.91 \\
    $M = 4$ & 33.36 & 0.954 & 0.049 & 129,664 & 2.81 \\
    \bottomrule
    \end{tabular}
    \label{tab:abl_GM}
\end{table}
We further conduct ablation studies on the proposed method, evaluated on \textit{cook spinach} and \textit{sear steak} data of N3DV dataset~\cite{n3dv}. We compare each hyperparameter of the Gaussian removal stage.  \cref{tab:abl_GP} and \cref{tab:abl_GM} show the comparison on different pruning ratio $p$ and the number of Gaussian Merging implementation, respectively. As expected for Gaussian-based models, the reconstruction quality increases monotonically with the number of Gaussians. We experimentally set these hyperparameters, considering the storage and image fidelity tradeoff. 

\subsection{Application on STG}
We further apply \textit{OMG4} to STG~\cite{spacetimegaussian}, to validate the generality of the proposed method. STG extends 3D Gaussian Splatting~\cite{3DGS} by incorporating temporal components to model dynamic scenes. In STG, the parameters of each Gaussian vary over time to capture object motions or appearance changes in the scene. Specifically, the spatial position of each Gaussian is modeled using a polynomial motion trajectory that predicts time-dependent positional displacements. For opacity and rotation, STG leverages a radial basis function and a polynomial function to capture temporal variations, respectively. In addition, to model appearance, spherical harmonics are replaced with neural features.

As shown in \cref{tab:stg}, \textit{OMG4} can achieve comparable reconstruction quality to CSTG~\cite{CSTG}, which is a compression technique designed for STG, while reducing the memory footprint by 27\%. Moreover, when using approximately 1.6 MB more memory than CSTG, \textit{OMG4} further improves the visual quality, demonstrating the scalability of our method. Combined with the results on Real-Time4DGS~\cite{realtime-4dgs} and FreeTimeGS~\cite{freetimegs}, these results verify that \textit{OMG4} generalizes well across different 4D Gaussian representation frameworks.

\begin{table}[t!]
    \centering
    \caption{Quantitative results of applying \textit{OMG4} to STG~\cite{spacetimegaussian} on N3DV~\cite{n3dv}.}
    \small
    \setlength{\tabcolsep}{4pt}
    \begin{tabular}{lccccc}
    \toprule
    Method & PSNR$\uparrow$ & SSIM$\uparrow$ & LPIPS (A)$\downarrow$ & \# G$\downarrow$ & Storage (MB)$\downarrow$ \\
    \midrule
    STG~\cite{spacetimegaussian} & 32.01 & 0.948 & 0.046 & 1,289,577 & 172 \\
    CSTG~\cite{CSTG} & 31.69 & 0.945 & 0.054 & -- & \cellcolor[HTML]{FFCCC9}15.4 \\
    \midrule
    Ours--L & 31.90 & 0.947 & 0.046 & \cellcolor[HTML]{FFCCC9}418,085 & 16.99 \\
    Ours--S & 31.72 & 0.945 & 0.050 & \cellcolor[HTML]{FFCE93}264,401 & \cellcolor[HTML]{FFCE93}11.17 \\
    \bottomrule
    \end{tabular}
    \label{tab:stg}
\end{table}

\subsection{Additional Results on N3DV}
\begin{table*}[t!]
    \centering
    \caption{Quantitative results on N3DV~\cite{n3dv}  at $1024 \times 768$ resolution. 
    Baseline results are sourced from the corresponding papers unless marked with $^*$; 
    $^*$ denotes reproduced results. 
    $^\bullet$ denotes results excluding the \textit{Coffee Martini} scene.}
    \resizebox{\linewidth}{!}{
    \begin{tabular}{lcccccc}
    \toprule
    Method & PSNR$\uparrow$ & SSIM$\uparrow$ & LPIPS (A)$\downarrow$ & FPS$\uparrow$ & \# G$\downarrow$ & Storage (MB)$\downarrow$ \\
    \midrule
    Real-Time4DGS~\cite{realtime-4dgs}$^*$ &32.21 & 0.950 & 0.040 & 96 & 2,770,350 & 1701 \\
    4DGS-1K~\cite{1000+} & 31.87 & 0.944 & 0.053 & 805 & 666,632 & 49.50\\
    Light4GS~\cite{Light4GS}$^\bullet$--L & 31.69 & - & 0.053 & 37 & - & 5.46  \\
    Light4GS~\cite{Light4GS}$^\bullet$--S & 31.48 & - & 0.064 & 40 & - & 3.77  \\
    \hline
    Ours--L & 32.21 & 0.948	& 0.044 & 162 &	231,537 &	4.77 \\
    Ours--M & 32.13 & 0.946 & 0.045 & 178 &\cellcolor[HTML]{FFFC9E}171,965 &\cellcolor[HTML]{FFFC9E}3.63 \\
    Ours--S & 32.08 & 0.945 & 0.048 & 185 &\cellcolor[HTML]{FFCCC9}136,679 &\cellcolor[HTML]{FFCCC9}2.95 \\
    Ours--T & 31.79 &	0.942 &	0.052 & 193 &\cellcolor[HTML]{FFCE93}98,217 &\cellcolor[HTML]{FFCE93}2.20 \\
    Ours$^\bullet$--L & 32.95 & 0.953 & 0.040 & 166 &217,401 &4.50 \\
    Ours$^\bullet$--T & 32.55 & 0.947 & 0.048 & 198 &92,033 & 2.08 \\
    \bottomrule
    \end{tabular}
    }
    \label{tab:quant_n3dv_1024}
\end{table*}

We additionally report N3DV results at $1024 \times 768$ resolution in~\cref{tab:quant_n3dv_1024}.\footnote{We report this additional comparison at $1024 \times 768$ because some baselines only provide results at this resolution.}
\textit{OMG4} achieves a strong rate-distortion trade-off, reducing 4DGS-1K~\cite{1000+} from 49.50 MB to 3.63 MB while improving PSNR (31.87 vs. 32.13). 
It also outperforms Light4GS--S~\cite{Light4GS} with less storage (31.48 vs. 32.13, 3.77 MB vs. 3.63 MB), confirming consistent compression gains.

\section{Analysis on Method}

\begin{figure}[t!]
\begin{center}
\includegraphics[width=1.0\linewidth]{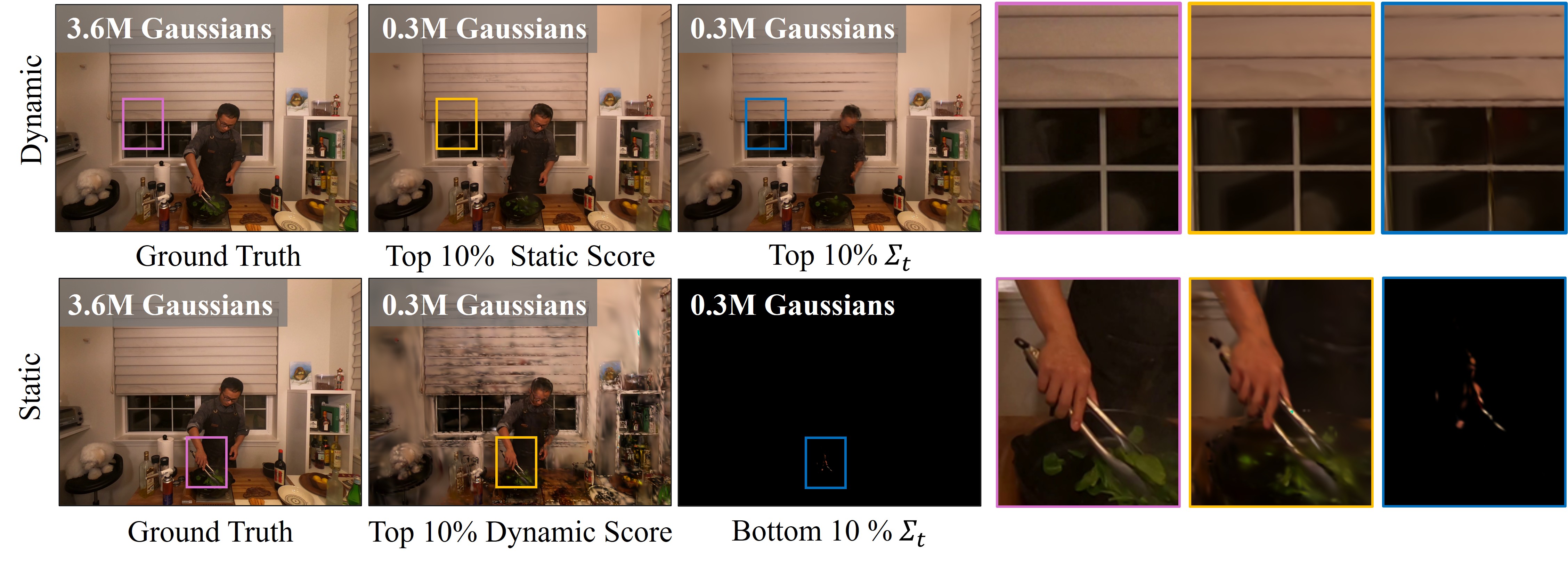}
\end{center}
   \caption{Effect of the sampling criterion under a fixed sampling budget. (Left) Rendered images with the full model. (Middle): Our $SD$ Score selection using top 10\% Static score and Dynamic Score. (Right) $\Sigma_t$-based sampling with the same sampling ratio.}
\label{fig:gaussian_sampling_compare}
\end{figure}

\begin{figure}[]
\begin{center}
\includegraphics[width=1.0\linewidth]{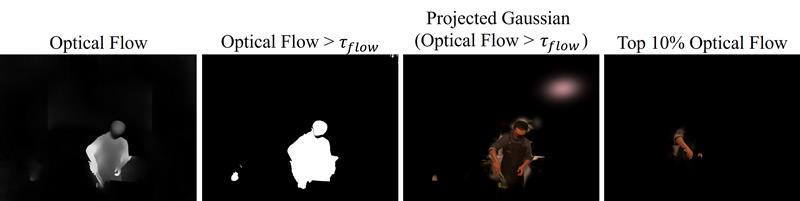}
\end{center}
   \caption{Visualization of 2D optical flow-based Dynamic score. (Left) Predicted 2D optical flow. (Middle-Left): Optical flow thresholded with $\tau_{flow}$. (Middle-Right) Rendered image with the Gaussians projected on high-optical flow region. (Right): Rendered image with the Gaussians of top 10\% Dynamic score.}
\label{fig:zeroshot}
\end{figure}

\begin{table}[t]
\centering
\caption{Ablation on replacing the proposed Dynamic Score with optical flow.}
\small
\setlength{\tabcolsep}{5pt}
\begin{tabular}{lccccc}
\toprule
Method & PSNR$\uparrow$ & SSIM$\uparrow$ & LPIPS$\downarrow$ & \# G$\downarrow$ & Storage (MB)$\downarrow$ \\
\midrule
Optical flow & 27.19 & 0.840 & 0.156 & 150,889 & 3.20 \\
Temporal gradient (Ours) &  31.75 & 0.939 & 0.062 & 135,684 & 2.93 \\
\bottomrule
\end{tabular}
\label{tab:zeroshot}
\end{table}

\subsection{SD Score}
In this section, we analyze the proposed $SD$ score in Gaussian Sampling stage. Gaussian Sampling aims to extract a subset of Gaussians that can faithfully represent both static and dynamic regions. Previous work~\cite{hybrid3d4d} has attempted to leverage the variance of time distribution, $\Sigma_t$, as a cue to detect separate static and dynamic regions. Gaussians located at the regions with various motions tend to have relatively smaller values of $\Sigma_t$ compared to those lying on the stationary area, as rapid changes in appearance or geometry restrict each primitive to have a short lifespan, being temporally localized. However, using $\Sigma_t$ alone to select a subset with a limited budget can be biased toward extremes because sampling dynamic Gaussians by taking the bottom $\Sigma_t$ subset focuses on a few high-velocity regions (e.g., moving hands in \cref{fig:gaussian_sampling_compare}), while neglecting other dynamic structures such as secondary body motion. This is because $\Sigma_t$ is agnostic to the visual contribution, oversampling transients that are temporally sharp. On the other hand, our $SD$ Score addresses this by combining gradient-based contribution with temporal support. Under the same sampling budget, it can distribute the budget across multiple dynamic regions while keeping enough static structures, as shown in~\cref{fig:gaussian_sampling_compare}, producing better perceptual fidelity with the same number of Gaussians.

Moreover, some online 4D reconstruction approaches~\cite{dai20254d, liu2024dynamics} adopts 2D optical flow to distinguish dynamic and static Gaussians using 2D optical flow. To further examine whether such approach can benefit our pipeline, we adopt the optical-flow--based dynamic region estimation proposed in~\cite{liu2024dynamics} and use it as a Dynamic Score. While streaming-based methods typically compute optical flow of each streamed frame, our method requires evaluating each Gaussian’s contribution to the entire scene globally. To this end, we accumulate optical-flow magnitudes across all frames. We then project each Gaussian onto the image plane and accumulate the flow magnitudes of pixels that fall within the corresponding 2D dynamic regions, thereby assigning an optical flow magnitude to each 4D Gaussian. If the accumulated optical flow exceeds a threshold $\tau_{flow}$, we mark such Gaussians as dynamic Gaussians. 

The segmented results are provided in \cref{fig:zeroshot} and quantitative results on the N3DV dataset are delivered in~\cref{tab:zeroshot}. The optical flow-based variant performs noticeably worse than our temporal gradient-based Dynamic Score in terms of both reconstruction quality and compression efficiency. We assume this is because optical flow is highly sensitive to frame-to-frame pixel changes and thus primarily highlights a few extremely fast-moving regions, rather than Gaussians that are essential for maintaining the global scene structure. As illustrated in \cref{fig:zeroshot}, the top 10\% Gaussians ranked by optical flow predominantly capture highly dynamic parts, but fail to select representative dynamic Gaussians necessary for preserving the overall scene, similar to using $\Sigma_t$~\cite{hybrid3d4d}. In other words, optical flow mainly measures how much a Gaussian move and does not guarantee sufficient coverage of the entire dynamic region under a tight sampling budget. In contrast, our temporal gradient-based Dynamic Score directly measures how much each Gaussian contributes to the reconstruction loss over time, leading to far more stable and reliable selections.

\subsection{Gaussian Merging}
\Cref{alg:gaussian_merging} describes a detailed process of Gaussian Merging. As mentioned earlier, our Gaussian Merging performs correlation-considered Gaussian removal based on the locality of Gaussians, while Gaussian Sampling and Gaussian Pruning rather aim to identify the salience or contribution of a single Gaussian. Also, computing proxy Gaussians ($\mathcal{P}_{\text{proxy}}$ in \cref{alg:gaussian_merging}) is differentiable, so the entire pipeline, including Gaussian Merging, can be trained end-to-end. It is worth to mention that once the weights ($w^x_i$ and $w^f_i$ in \cref{alg:gaussian_merging}) are learned, Gaussians in each clusters are permanently merged via weighted sum; therefore, relevant training-time overhead disappears during the rest of the process (\textit{i.e.,} Attribute Compression). At test time, since the Gaussians are already merged, the merging operation incurs no extra cost and does not affect FPS.

\begin{algorithm}[]
\caption{Gaussian Merging}
\label{alg:gaussian_merging}
\begin{algorithmic}[0]
\Require Gaussian set $\mathcal{P}_{GP}$;
clusters $\mathcal{C}=\{C_q\}_{q=1}^{N_C}$;
training steps $T_M$
\State Initialize per-member logits $\ell^x_i\!\leftarrow\!0,\ \ell^f_i\!\leftarrow\!0$ for all $i$
\For{$t=1$ {\bf to} $T_M$}
  \ForAll{clusters $C_q$}
    \State $w^x_i \gets \sigma(\ell^x_i)\big/\sum_{j\in C_q}\sigma(\ell^x_j)$ \quad for $i\in C_q$ \Comment{Position weights}
    \State $w^f_i \gets \sigma(\ell^f_i)\big/\sum_{j\in C_q}\sigma(\ell^f_j)$ \quad for $i\in C_q$ \Comment{Appearance weights}
    \State $\bar{{x}}_q \gets \sum_{i\in C_q} w^x_i\,{x}_i$;\quad
           $\bar{{f}}_q \gets \sum_{i\in C_q} w^f_i\,{f}_i$
    \State $\bar{{a}}_q \gets {a}_{r(C_q)}, \quad  r(C_q)= \arg\max_{i \in C_q} {w}_{i}^{x}$
  \EndFor
  \State $\mathcal{P}_{\text{proxy}} \gets \{(\bar{{x}}_q,\bar{{f}}_q,\bar{{a}}_q)\}_{q=1}^{N_C}\ \cup\ \{({x}_i,{f}_i,{a}_i): i\notin \bigcup_q C_q\}$
  \State Render $\mathcal{P}_{\text{proxy}}$, compute loss $\mathcal{L}$, and update $\{\ell_i^x,\ell_i^f\}$ by backprop
\EndFor
\State \Return pruned set $\mathcal{P}_{\text{proxy}}$
\end{algorithmic}
\end{algorithm}

\subsection{Training Time Analysis}

\cref{tab:time} reports the optimization time of \textit{OMG4} applied to Real-Time4DGS~\cite{realtime-4dgs} on the N3DV dataset~\cite{n3dv}, measured with an RTX 4090 GPU. We train \textit{OMG4} for a total of 11,000 iterations, and, before optimization, we perform a preprocessing step that computes view-space and temporal gradients. The row “OMG4” therefore includes both this preprocessing and the subsequent optimization, while “OMG4*” excludes the gradient computation and reports only the pure optimization time. Thanks to our effective reduction of the Gaussian count, the per-iteration optimization time decreases from 0.31 s/iter for the original pretraining stage to 0.157 s/iter for OMG4*, i.e., nearly a 2$\times$ speedup. The lower part of the table further decomposes the wall-clock time of each module. “Gradient Computing” corresponds to the one-time cost of preprocessing.

Both “Gaussian Sampling” and “Gaussian Pruning” run for 1,000 iterations. “Gaussian Pruning” is applied after “Gaussian Sampling” has already reduced the number of points, which explains its slightly lower per-iteration time. For “Gaussian Merging”, we separately list the time for cluster computation (“Calc. Clusters”) and the subsequent optimization. During merging, Gaussians are temporarily combined for rendering, which introduces a small overhead of about 0.01 second per iteration; in the second merging stage, the number of Gaussians is already reduced, so both clustering and optimization are faster than in the first stage.

In the “Attribute Compression” block, we show that using a small MLP for attribute encoding has a negligible impact on runtime and that the additional stages with SVQ-3D and SVQ-4D introduce only a modest increase in per-iteration time. After optimization is done, decoding the compressed model takes approximately 8 seconds, which is a one-time cost. Overall, these results demonstrate that our pipeline substantially reduces optimization time per iteration while keeping each compression and merging component computationally lightweight.

\begin{table*}[t]
    \centering
    \caption{Training time analysis.}
    \small
    \setlength{\tabcolsep}{6pt}
    \begin{tabular}{llccc}
    \toprule
    Method & Sub-module & \# Iterations & Time (min) & Time / iter (sec) \\
    \midrule
    Pretraining & -- & 30,000 & 155.05 & 0.310 \\
    \textit{OMG4} & -- & 11,000 & 45.17 & 0.246 \\
    \textit{OMG4}$^*$ & -- & 11,000 & 28.83 & 0.157 \\
    \midrule
    Gradient Computing & -- & -- & 16.35 & -- \\
    Gaussian Sampling & -- & 1,000 & 2.47 & 0.148 \\
    Gaussian Pruning & -- & 1,000 & 2.41 & 0.144 \\
    Gaussian Merging 1 & Cluster calculation & -- & 0.83 & -- \\
    Gaussian Merging 1 & Optimization & 1,000 & 2.57 & 0.154 \\
    Gaussian Merging 2 & Cluster calculation & -- & 0.59 & -- \\
    Gaussian Merging 2 & Optimization & 1,000 & 2.44 & 0.146 \\
    Attribute Compression & MLP & 5,000 & 12.14 & 0.146 \\
    Attribute Compression & MLP + 3D SVQ & 1,000 & 2.57 & 0.154 \\
    Attribute Compression & MLP + 3D SVQ + 4D SVQ & 1,000 & 2.83 & 0.170 \\
    \bottomrule
    \end{tabular}
    \label{tab:time}
\end{table*}

\section{Additional Implementation Details}

In this section, we describe the implementation details and experimental settings in detail. 
\subsection{Real-Time4DGS setup} As mentioned earlier, we follow the Sub-Vector Quantization (SVQ) configuration of OMG~\cite{OMG}, where the codebook sizes for scale $b_s$, rotation $b_r$, and encoded appearance feature $b_f$ are set to $2^9$, $2^{13}$, and $2^{10}$, respectively. Since Real-Time4DGS~\cite{realtime-4dgs} introduces additional attributes, namely $s_t$ (the scale along the temporal axis) and $q^l$ (an additional quaternion used to define the rotation matrix), we also apply SVQ to these attributes, using codebook sizes of $2^9$ and $2^{13}$, respectively. We use this configuration for all datasets and all variants. We use $\tau_{GS}$ as 0.2, 0.3, 0.2 and quantile $p$ as 0.7, 0.9, 0.9 for our large, small and tiny variants, respectively. For the ENeRF-Outdoor~\cite{enerf} dataset, we set $\tau_{GS}=0.2$ and use the 0.9-quantile.

\subsection{Free-TimeGS setup} Similar to the Real-Time4DGS~\cite{realtime-4dgs} setting, we also apply sub-vector quantization to the additional attributes introduced by FreeTimeGS, namely velocity and duration. We use the same codebook sizes for both model scales, setting the codebook size of both velocity and duration to $2^{10}$.
For the N3DV~\cite{n3dv} dataset, we set $\tau_{GS}$ to 0.5, 0.7, 0.5, 0.4 and quantile $p$ as 0.6, 0.8, 0.8, 0.8 for our large, medium, small, and tiny variants and set $\tau_{GS}$ to 0.4, 0.4, 0.3, 0.3 and quantile $p$ as 0.2, 0.4, 0.4, 0.6 for our large, medium, small, and tiny variants in SelfCap~\cite{selfcap} dataset. For the ENeRF-Outdoor~\cite{enerf} datasets, we set $\tau_{GS}=0.5$ and use the 0.2-quantile for Gaussian Pruning.

As shown in the ablation studies provided earlier, we found that the reconstruction quality generally improves as more Gaussians are preserved. This scalability allows our method to flexibly control the rate-distortion trade-off, making it adaptable to different memory budgets and practical requirements.

\section{Qualitative Comparisons}
\begin{figure}[t!]
\begin{center}
\includegraphics[width=1.0\linewidth]{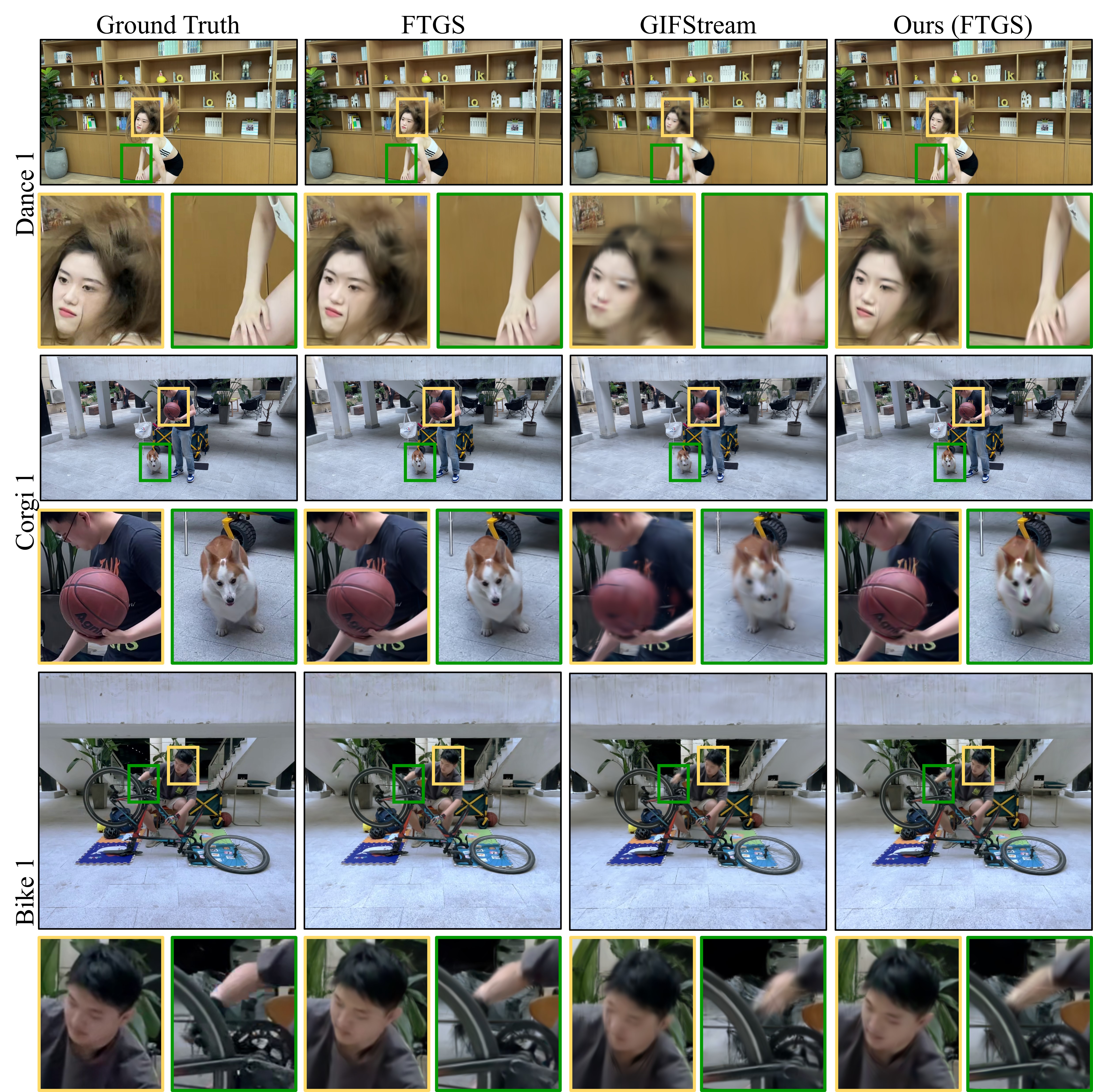}
\end{center}
   \caption{Qualitative results on SelfCap dataset~\cite{selfcap}. }
\label{fig:selfcap_qualitative}
\end{figure}

\begin{figure}[t!]
\begin{center}
\includegraphics[width=1.0\linewidth]{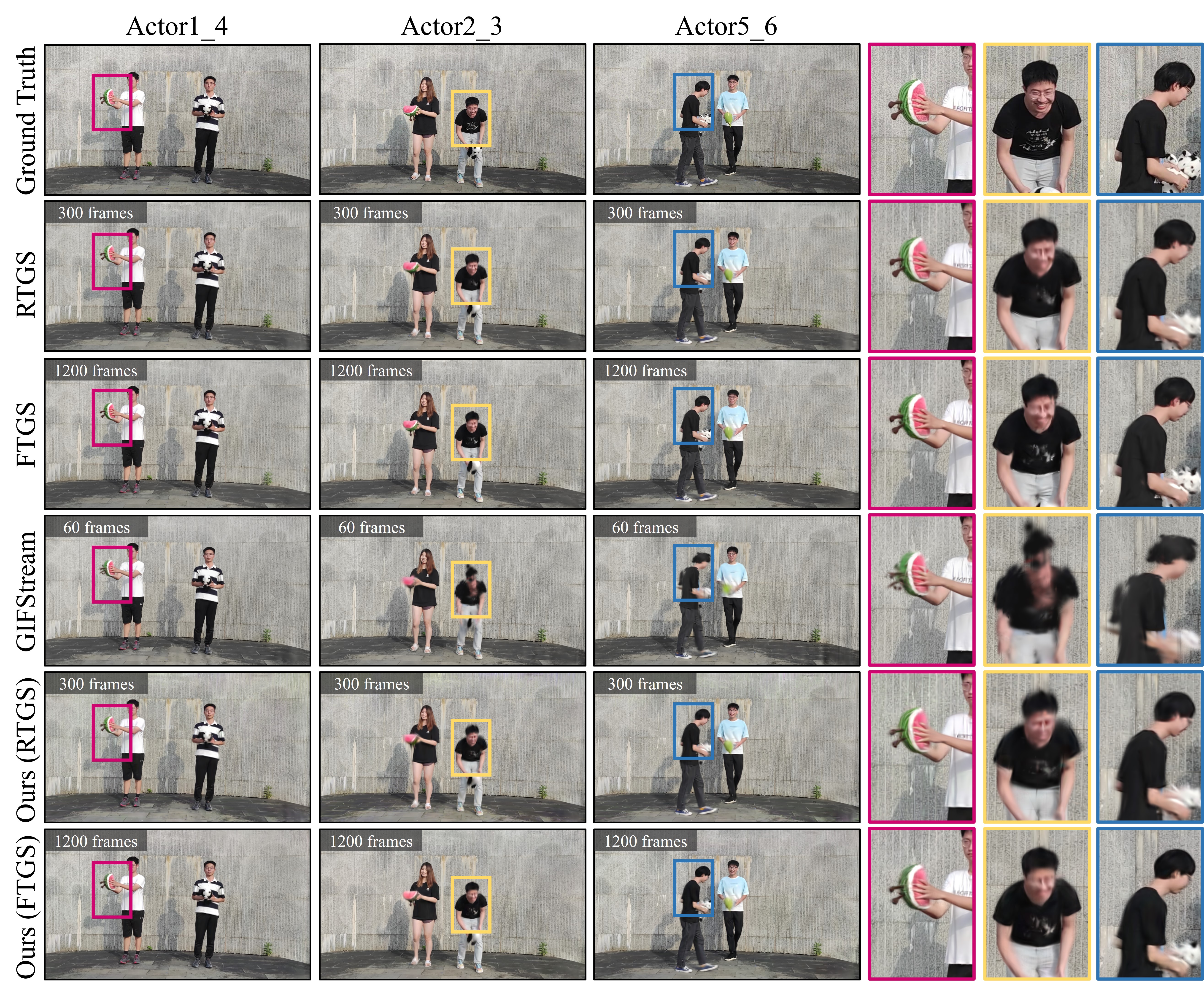}
\end{center}
   \caption{Qualitative results on ENeRF-Outdoor dataset~\cite{enerf}.}
\label{fig:enerf_qualitative}
\end{figure}

We provide visual comparisons of SelfCap~\cite{selfcap} and ENeRF-Outdoor~\cite{enerf} dataset in \cref{fig:selfcap_qualitative} and \cref{fig:enerf_qualitative}.

\clearpage

\clearpage


\newpage

\end{document}